\documentclass[11pt,a4paper]{article}

\usepackage[T1]{fontenc}
\usepackage[utf8]{inputenc}
\usepackage{geometry}
\usepackage{amsmath,amssymb,amsthm}
\usepackage{booktabs}
\usepackage{multirow}
\usepackage{tabularx}
\usepackage{longtable}
\usepackage{array}
\usepackage{makecell}
\usepackage{xcolor}
\usepackage{colortbl}
\usepackage{graphicx}
\usepackage{adjustbox}
\usepackage{subcaption}
\usepackage{float}
\usepackage{soul}
\usepackage{algorithm}
\usepackage{algpseudocode}
\usepackage{enumitem}
\usepackage{setspace}
\usepackage{fancyhdr}
\usepackage{titlesec}
\usepackage{natbib}
\usepackage{mdframed}
\usepackage{url}
\usepackage{hyperref}

\newcommand{\frameworkname}{BADGER}
\newcommand{\orgname}{Merkle}

\definecolor{merklered}{RGB}{196,0,45}
\definecolor{lightgray}{RGB}{245,245,245}
\definecolor{pendingeditcolor}{RGB}{0,90,190}
\newcommand{\pendingedit}[1]{#1}

\hypersetup{
  colorlinks=true,
  linkcolor=black,
  citecolor=black,
  urlcolor=black,
  pdftitle={BADGER: Bridging Agentic and Deterministic evaluation for Generative Enterprise Reasoning},
  pdfauthor={Shannon Serrao, Soumitra Chatterjee, Dorina Strori, Abhishek Sharma, Nathan Miller}
}

\title{
  \vspace{-1cm}
  {\Large \textbf{\frameworkname: Bridging Agentic and Deterministic evaluation for Generative Enterprise Reasoning}}
  \vspace{4pt}
  
  {\normalsize \textit{A Unified Evaluation Framework for Enterprise Agentic AI Conversational Systems}}
}

\author{
  Shannon Serrao\thanks{Merkle Analytics.
    Correspondence: \texttt{shannon.serrao@merkle.com}}\quad
  Soumitra Chatterjee$^*$\quad
  Dorina Strori$^*$ \\[4pt]
  Abhishek Sharma$^*$\quad
  Nathan Miller$^*$ \\[6pt]
  {\normalsize\textit{Merkle Analytics}}
}

\date{May 2026}

\begin{document}
\sloppy
\maketitle
\thispagestyle{fancy}

\begin{abstract}
Enterprise conversational insights AI systems that translate natural language into SQL queries against production data warehouses and increasingly orchestrate this capability within multi-step agentic reasoning pipelines, require evaluation approaches fundamentally different from those offered by academic benchmarks. Spider and BIRD established
rigorous execution-accuracy protocols on controlled schemas; G-Eval and
RAGAS advanced LLM-based quality assessment; and recent work such as
Spider~2.0 \citep{lei2024spider2}, BEAVER \citep{chen2024beaver}, and
BIRD-Interact \citep{birdinteract2025} has begun to address enterprise
and agentic dimensions. Nevertheless, no single framework unifies
text-to-SQL assessment with agentic behavior evaluation into a
production-grade pipeline calibrated against human expert judgment.

We present \frameworkname{}, developed at \orgname{}, for deployment within governed enterprise data environments, as a unified evaluation framework that coherently integrates text-to-SQL assessment with agentic behavior evaluation. \frameworkname{} offers three primary contributions.
First, it introduces an LLM-assisted SQL component extraction that extends
the Spider methodology to handle the CTE-heavy, dialect-specific SQL
produced by modern LLMs. Second, it introduces a \emph{hybrid execution
accuracy} metric, our primary novel contribution, that
resolves the column-aliasing and numeric-tolerance brittleness of standard
execution accuracy by using an LLM to infer structural alignments before
applying fully deterministic cell-level scoring. This approach is
empirically validated on 150 human-annotated industry queries, achieving
Cohen's $\kappa = 0.717$ [95\% CI: 0.600--0.822] \emph{Substantial}
agreement per Landis \& Koch \citeyearpar{landis1977} and a balanced
accuracy of 87.3\%, outperforming all six competing evaluation frameworks
on human alignment ($\Delta\kappa$ range: 0.322--0.502, all $p \leq
0.001$). Third, it incorporates an enterprise agentic evaluation suite
that assembles established evaluation metrics from prior frameworks---RAGAS
for response faithfulness \citep{es2023ragas}, G-Eval for summary quality
\citep{liu2023geval}, and prior agent benchmark work for tool-call
evaluation \citep{patil2023gorilla,liu2023agentbench}---into a unified
production pipeline. The Excess Tool Usage metric is the sole novel agentic
element; the suite's contribution is the integration of these metrics into
a production-grade pipeline rather than the individual metrics themselves.

\frameworkname{} is delivered as an
internal evaluation tool that runs entirely within the client's
governed data environment, enabling Merkle clients across financial
services, retail, and healthcare to submit agent runs and receive
structured evaluation reports. The tool supports configurable LLM
judge backends and the rapid prototyping of production-grade,
client-specific judges and metrics that complement the standard suite,
allowing tailored evaluation criteria to be built and tested alongside the
core methodology. By addressing the evaluation failure modes that matter most in enterprise data delivery (column-aliasing false negatives, numeric-tolerance brittleness, and the gap between agentic tool orchestration and response quality), BADGER is designed to serve as a continuous evaluation backbone for production AI systems rather than a one-time quality gate. 
\end{abstract}

\newpage
\tableofcontents
\newpage

\section{Introduction}
\label{sec:intro}

The aspiration of enabling non-technical business users to query
enterprise data warehouses in plain English has driven decades of
research \citep{zelle1996, popescu2003}. The arrival of large language
models has made this largely achievable: state-of-the-art systems now
exceed 90\% execution accuracy on controlled benchmarks
\citep{gao2023,pourreza2023dinsql}. Yet practitioners deploying these
systems report a persistent gap between benchmark scores and real-world
reliability, a gap documented empirically by \citet{floratou2024nl2sql}
from an industry perspective.

This gap is partly technical and partly methodological. Enterprise SQL
environments differ from research benchmarks in ways that matter:
production databases use cloud-native dialects (Snowflake, BigQuery,
Databricks SQL) with constructs absent from benchmark training corpora;
business questions are compound, ambiguous, and sometimes unanswerable;
and correct answers may require joining across many tables under evolving
schemas. On the methodological side, the evaluation tools provided by
the research community do not scale to the continuous evaluation needs of
a live enterprise product.

The challenge has deepened with the shift to agentic architectures. A
modern enterprise data agent does not simply map a question to a SQL
query; it decomposes the question, selects from a palette of tools (SQL
generation, semantic search, metadata lookup, summarization), executes a
chain of reasoning steps, and produces a natural-language response
grounded in retrieved data \citep{yao2022react, wang2023plansolve}. No
existing framework addresses both dimensions in a unified pipeline
calibrated against human expert judgment.

This paper presents \frameworkname{}, a unified evaluation framework
developed at \orgname{} to fill this gap. We make the following
contributions:

\begin{enumerate}[leftmargin=*]
\item \textbf{LLM-assisted SQL component extraction} extending
  Spider-style component matching to modern LLM-generated SQL without
  requiring a deterministic parser (\S\ref{sec:component_extraction}).
\item \textbf{A four-tier SQL complexity classifier} for enterprise
  LLM-generated SQL, adopting Spider's label scheme with a
  redesigned signal set that adds CTE detection
  (\S\ref{sec:complexity}).
\item \textbf{An LLM judge for SQL quality} with an explicit issue
  taxonomy and a carefully designed rubric (\S\ref{sec:llm_judge}).
\item \textbf{Hybrid execution accuracy (Hybrid-EX)} (our primary
  novel contribution), a two-stage metric that uses an LLM to resolve
  structural ambiguities (column aliasing, table mapping) before applying
  a fully deterministic cell-level scorer, achieving Substantial human
  agreement ($\kappa = 0.717$ [95\% CI: 0.600--0.822]) and
  significantly outperforming all six evaluated baselines ($\Delta\kappa$
  range: 0.322--0.502, all $p \leq 0.001$) (\S\ref{sec:hybrid_ex}).
\item \textbf{Empirical validation} against human expert labels across
  150 human-annotated industry queries, with statistically rigorous
  comparisons against six competing frameworks using bootstrap pairwise
  Cohen's $\kappa$ tests (\S\ref{sec:validation}).
\item \textbf{An enterprise agentic evaluation suite} assembling
  established metrics from existing frameworks---RAGAS
  \citep{es2023ragas}, G-Eval \citep{liu2023geval}, and prior agent
  benchmark work \citep{patil2023gorilla,liu2023agentbench}---into a
  unified production pipeline, with the Excess Tool Usage metric as the
  sole novel element (\S\ref{sec:agentic}). These metrics are included
  for production evaluation completeness; they do not represent
  additional novel methodological contributions of this paper.
\item \textbf{A production deployment} that runs within the client's
  governed data environment, supporting developer and client-facing
  evaluation runs with configurable LLM judge backends, custom scoring
  rubrics, and the rapid prototyping of production-grade,
  client-specific judges and metrics alongside the standard suite
  (\S\ref{sec:deployment}).
\end{enumerate}

\section{Background and Related Work}
\label{sec:related}

\subsection{The Text-to-SQL Evaluation Landscape}

\paragraph{Early benchmarks and execution accuracy.}
Text-to-SQL evaluation was formalized at scale by WikiSQL
\citep{zhong2017seq2sql}, which introduced execution accuracy
(EX), the test of whether the generated query returns the same result set as the
gold query, as the primary metric. Spider \citep{yu2018spider} introduced
two critical advances: a cross-domain benchmark spanning 200 databases,
and a component-level exact match (EM) that compares SQL queries at the
structural level. Spider's four-tier hardness taxonomy stratified results
by complexity.

\paragraph{BIRD and realistic execution.}
BIRD \citep{li2024bird} demonstrated that execution accuracy on clean
research databases overestimates real-world performance, introducing
``dirty'' databases and evidence-augmented evaluation.
\citet{zhong2020semantic} introduced test suite accuracy by generating
multiple test databases to evaluate semantic equivalence, an approach directly
relevant to the execution accuracy limitations \frameworkname{} addresses.

\paragraph{Enterprise and dialect-aware benchmarks.}
The most directly relevant recent work is \textbf{Spider~2.0}
\citep{lei2024spider2}, which evaluates enterprise-level text-to-SQL with
multi-dialect SQL (BigQuery, Snowflake), large-scale schemas (3,000+
columns), and agentic coding workflows. \textbf{BEAVER}
\citep{chen2024beaver} is the first enterprise benchmark sourced from
real private data warehouses with actual query logs, directly validating
\frameworkname{}'s motivating claim about the gap between academic benchmarks
and enterprise deployment. \textbf{KaggleDBQA} \citep{lee2021kaggle}
and \textbf{Dr.~Spider} \citep{chang2023drspider} expose brittleness
through robustness perturbations; \citet{floratou2024nl2sql} provides
industry evidence that benchmark scores systematically overstate
production performance.
\citet{sun2021mtteql} introduced metamorphic testing for text-to-SQL
consistency evaluation, generating semantically equivalent NL variants to
expose inconsistencies in model outputs, a robustness angle complementary
to \frameworkname{}'s production-calibration focus.
\textbf{NL2SQL360} \citep{li2024nl2sql360} provides a multi-angle
evaluation framework with fine-grained performance breakdowns by SQL
characteristic (joins, subqueries, aggregations) and application domain,
offering a complementary perspective to \frameworkname{}'s human-calibration
approach.

\paragraph{UNITE and prior unification attempts.}
\textbf{UNITE} \citep{lan2023unite} represents a prior attempt at
unification across text-to-SQL benchmarks from AWS, directly relevant to
\frameworkname{}'s unified framing and from which it should be
differentiated: UNITE aggregates benchmarks for training, while
\frameworkname{} focuses on production evaluation with human calibration.

\paragraph{LLM-era Text-to-SQL.}
\citet{gao2023} demonstrated that chain-of-thought decomposition
significantly improves LLM text-to-SQL accuracy.
\citet{pourreza2023dinsql} showed that decomposed in-context learning
with self-correction reduces errors. These methods produce SQL with
stylistic patterns (deeply nested CTEs, positional GROUP BY,
schema-qualified aliases) that deterministic component-matching parsers
mishandle.

\paragraph{LLM-based execution accuracy.}
\citet{kim2025flex} introduced FLEX (False-Less EXecution), an
LLM-based metric that uses a strong LLM judge, given the natural
language question, database schema, both SQL queries, and both
execution results, to determine semantic equivalence between gold
and candidate queries. FLEX reports Cohen's $\kappa = 0.87$ on
the Spider and BIRD academic benchmarks against expert human
labels, establishing the strongest published reference point for
LLM emulation of expert SQL judgment.

FLEX is not included as a head-to-head baseline because its
published implementation is a Spider/BIRD benchmark runner rather
than a general-purpose evaluation library, and a faithful
enterprise run would require non-trivial adapter work whose
output would no longer be directly comparable to FLEX as
published.\footnote{The FLEX codebase (Apache 2.0) validates the
dataset argument against a fixed set of identifiers, its schema
loader expects Spider's \texttt{dev\_tables.json} format, and its
judge prompt was calibrated on Spider/BIRD schemas. A faithful
enterprise run would require forking the codebase, writing a
schema-format adapter from cloud-warehouse catalogs to FLEX's
expected JSON, and re-formatting evaluation records into FLEX's
BIRD-style input layout. The resulting score would characterize
the adaptation, not FLEX itself; reporting it against Kim et
al.'s $\kappa = 0.87$ without a separate calibration study to
verify the adapter's faithfulness would risk misrepresenting
either their contribution or our own results.} We therefore
treat FLEX as a strong academic-benchmark reference for the
LLM-emulation paradigm rather than as a directly competing
implementation.

\frameworkname{}'s Hybrid-EX differs from both FLEX and LLM-EX in
three respects: it isolates LLM involvement to structural
alignment inference only, applying a fully deterministic scoring
stage that eliminates run-to-run scoring variance; it explicitly
handles the column-aliasing and numeric-tolerance failure modes
common in enterprise LLM-generated SQL; and it is validated on
enterprise queries rather than academic benchmarks. The two
approaches are therefore complementary: FLEX prioritizes LLM
emulation of expert judgment on academic benchmarks, while
Hybrid-EX prioritizes reproducibility and structured error
decomposition in a production pipeline.

\subsection{LLM-Based Evaluation}

\pendingedit{\paragraph{LLM-as-judge and G-Eval.}
\citet{zheng2023judging} demonstrated that LLMs can serve as reliable
evaluators of other LLMs' outputs. \textbf{G-Eval} \citep{liu2023geval}
formalized LLM-based evaluation of NLG quality using a
chain-of-thought (CoT) form-filling paradigm: an LLM is given a task
description and evaluation criteria, auto-generates intermediate
evaluation steps via CoT, then scores the candidate output on each
dimension. Final scores are computed as a probability-weighted
summation of output tokens, yielding finer-grained continuous
scores than direct integer prediction.
G-Eval with GPT-4 achieves a Spearman correlation of 0.514 with
human judgments on the SummEval summarization benchmark
\citep{fabbri2021summeval}, outperforming all prior automatic
evaluators across Coherence, Consistency, Fluency, and
Relevance, the same four dimensions \frameworkname{} adopts.
\textbf{FLASK} \citep{ye2024flask} addresses fine-grained LLM
evaluation with rubrics and documents biases in LLM-as-judge
methodologies, biases that \frameworkname{}'s hybrid design
explicitly mitigates by using the LLM only for structural
inference, not for final scoring.}

\paragraph{Faithfulness in RAG systems.}
RAGAS \citep{es2023ragas} formalized faithfulness evaluation for
retrieval-augmented generation. In the enterprise data agent context, the
analogous failure is a summary that fabricates trends or percentages not
present in the SQL result table.

\subsection{Agentic AI Evaluation}

\paragraph{Tool-use evaluation.}
\citet{patil2023gorilla} evaluated LLMs on their ability to select and
invoke API functions, documenting hallucinated tool calls. This failure
mode translates directly to enterprise data agents.

\paragraph{Agentic Text-to-SQL.}
\textbf{BIRD-Interact} \citep{birdinteract2025}, accepted at ICLR 2026,
introduces conversational (c-Interact) and agentic (a-Interact)
interaction modes for text-to-SQL evaluation, including Interaction-Time
Scaling. This is the most directly competing work to \frameworkname{}'s
unified claim; the key differentiator is that \frameworkname{} provides
production-calibrated evaluation deployed within a governed enterprise
data environment, while BIRD-Interact remains a research benchmark.

\paragraph{Agent benchmarks.}
AgentBench \citep{liu2023agentbench} provides the most comprehensive
general-purpose benchmark for LLM agents but does not evaluate SQL
generation quality, response faithfulness, or multi-intent decomposition.
ReAct \citep{yao2022react} and Reflexion \citep{shinn2023reflexion}
contribute prompting strategies but not evaluation frameworks.
\citet{deepeval2024agents} catalogs purpose-built agent evaluation
metrics for tool use, task completion, and reasoning chains; \frameworkname{}'s
agentic suite addresses several of these dimensions in the enterprise
text-to-SQL context specifically.

\paragraph{Conversational and multi-turn evaluation.}
SParC \citep{yu2019sparc} and CoSQL \citep{yu2019cosql} address
multi-turn conversational text-to-SQL, directly relevant to
\frameworkname{}'s intent resolution and clarification evaluation metrics.

\subsection{Summary: Gaps Addressed by \frameworkname{}}

Table~\ref{tab:coverage} positions \frameworkname{} relative to prior
work on the text-to-SQL evaluation dimensions that represent this
paper's novel contributions: LLM-dialect robustness in component
extraction, numeric/alias tolerance in execution accuracy validated
against human labels, and a unified production pipeline with
configurable backends.

\begin{table}[htbp]
\centering

\caption{Text-to-SQL evaluation capabilities: \frameworkname{} vs.\ prior work.
  $\checkmark$~=~directly addressed; $\sim$~=~partially addressed;
  --~=~not addressed. This table covers the text-to-SQL dimensions that
  constitute \frameworkname{}'s novel methodological contributions.
  Agentic metrics (faithfulness, G-Eval, tool-call evaluation) are
  assembled from existing frameworks for production completeness and are
  not included in this comparison. FLEX \citep{kim2025flex} is omitted
  because it is a benchmark-specific runner rather than a general-purpose
  evaluation library; see \S\ref{sec:related} for discussion.}
\label{tab:coverage}
\resizebox{\linewidth}{!}{%
\footnotesize
\setlength{\tabcolsep}{4pt}
\begin{tabular}{lccccc}
\toprule
\textbf{Capability}
  & \textbf{Spider} & \textbf{BIRD} & \textbf{Spider 2.0}
  & \textbf{BIRD-Int.} & \textbf{\frameworkname{}} \\
\midrule
SQL component matching  & $\checkmark$ & $\checkmark$ & $\sim$       & --           & $\checkmark$ \\
Execution accuracy      & $\checkmark$ & $\checkmark$ & $\checkmark$ & $\checkmark$ & $\checkmark$ \\
LLM dialect robustness  & --           & $\sim$       & $\checkmark$ & $\sim$       & $\checkmark$ \\
Numeric/alias tolerance & --           & --           & --           & --           & $\checkmark$ \\
Human-label calibration & --           & --           & --           & --           & $\checkmark$ \\
SQL qualitative judge   & --           & --           & --           & --           & $\checkmark$ \\
Query complexity tiers  & $\sim$       & --           & $\sim$       & --           & $\checkmark$ \\
Governed enterprise deployment & --    & --           & --           & --           & $\checkmark$ \\
Custom LLM judge backends      & --    & --           & --           & --           & $\checkmark$ \\
Rapid custom metric prototyping& --    & --           & --           & --           & $\checkmark$ \\
Unified production pipeline    & --    & --           & --           & --           & $\checkmark$ \\
\bottomrule
\end{tabular}%
}

\vspace{6pt}
\begin{minipage}{\textwidth}
  \footnotesize\textit{Coverage gaps.}\enspace
  Two evaluation dimensions are absent from all frameworks above
  and represent planned extensions to \frameworkname{}.
  \emph{Multi-turn conversational evaluation}, which assesses follow-up
  questions, pronoun coreference, and session-level context in the
  style of SParC \citep{yu2019sparc} and CoSQL
  \citep{yu2019cosql}, is not yet part of the standard suite;
  intent resolution (\S\ref{sec:agentic}) covers single-turn
  decomposition and clarification but not full conversational chains.
  \emph{Multimodal RAG evaluation}, which verifies that chart and
  visualization outputs faithfully reflect the underlying retrieved
  data, is currently under development (see \S\ref{sec:deployment}).
  Both are discussed further in \S\ref{sec:future}.
\end{minipage}

\end{table}

\section{Framework Design}
\label{sec:design}

\subsection{Design Principles}

\frameworkname{} was designed around four principles derived from
production experience:

\paragraph{Holistic coverage.} No single metric tells the whole story.
\frameworkname{} measures all relevant dimensions simultaneously, enabling practitioners to understand not just whether a system fails but how and where. The standard metric suite (Table 2) provides production-ready evaluation out of the box, but enterprise deployments invariably surface failure modes unique to a specific client, vertical, or use case. A financial services deployment may require regulatory compliance checks that a retail deployment does not, while a healthcare deployment may need domain-specific factual guardrails absent from any public benchmark. BADGER  addresses this through a custom judge interface (\S\ref{sec:config}) that allows practitioners to build, validate, and deploy client-specific judges and metrics alongside the standard suite without modifying the core pipeline. The standard metrics represent our methodological contribution; the customization framework ensures that holistic coverage extends to the long tail of evaluation needs that no fixed metric set can anticipate. 

\paragraph{Determinism where possible, LLM where necessary.} LLM-based
metrics introduce non-determinism and cost. \frameworkname{} uses LLMs only
for tasks requiring semantic understanding (component extraction,
qualitative judgment, structural alignment inference) and applies fully
deterministic algorithms for all quantitative scoring once the semantic
context is resolved. The hybrid execution accuracy (Hybrid-EX) metric embodies
this principle directly: once the LLM-based table-mapping stage fixes the
structural alignment, the scoring stage is fully deterministic,
eliminating run-to-run score variance by construction.

\paragraph{Calibration to human expert judgment.} Unlike academic
benchmarks evaluated solely on internal consistency, \frameworkname{}'s
execution accuracy approach is calibrated against 150 human-annotated
industry queries labeled by three independent expert annotators,
achieving Cohen's $\kappa = 0.717$ [95\% CI: 0.600--0.822]
(\S\ref{sec:validation}). Human ground truth quality is itself
validated: Fleiss' $\kappa = 0.746$ and Krippendorff's $\alpha = 0.749$
\citep{fleiss1971, krippendorff2004}, with 82.0\% of cases unanimous. The current corpus size reflects the constraints of validating on real enterprise data: every query in the evaluation set was drawn from live client deployments and required data governance approval before inclusion in the annotation study. This constraint, which is inherent to any framework that validates on production data rather than public benchmarks, limits the initial sample size but ensures that the calibration reflects genuine enterprise conditions rather than synthetic approximations. As additional client engagements complete governance review, the annotated corpus will be expanded to narrow the confidence intervals reported in (\S\ref{sec:roc_auc}). For the same reason, the evaluation corpus is not publicly released. The methodological contribution, including all algorithms, metric definitions, prompt specifications, and the evaluation protocol, is fully specified in this paper and can be applied by practitioners to their own data.

\paragraph{Graceful degradation and longitudinal comparability.}
Real evaluation pipelines encounter missing data, failed executions, and
LLM API timeouts. \frameworkname{} is designed so that any single failure
affects only the metrics for the failing row; aggregate statistics remain
valid, and failure reasons are recorded for debugging. Every evaluation
run carries stable metadata enabling longitudinal tracking across agent
versions.

\subsection{Evaluation Architecture}

\frameworkname{} operates on a golden test set prepared offline. Each record
pairs a natural-language question with: the expert-authored gold SQL
query and its pre-executed result table; the expected sequence of tool
calls for the agentic branch; and, for ambiguous queries, a set of
expected clarification questions.

The framework supports three evaluation modes:

\begin{itemize}[leftmargin=*]
\item \texttt{TEXT2SQL\_ONLY}   Activates the SQL evaluation branch.
\item \texttt{AGENTIC\_ONLY}   Activates the agentic branch.
\item \texttt{BOTH}   Standard mode for enterprise data agents.
\end{itemize}

The complete metric inventory is given in Table~\ref{tab:metrics}.

\begin{table}[htbp]
\centering
\caption{\frameworkname{} complete metric inventory.}
\label{tab:metrics}
\footnotesize
\begin{tabular}{p{1.8cm}p{4.2cm}p{7.5cm}}
\toprule
\textbf{Branch} & \textbf{Metric} & \textbf{Description} \\
\midrule
\multirow{14}{*}{\makecell{Text-to-\\SQL}} &
  Syntactic Accuracy & Binary: SQL executed without error. \\
& Component Recall (7 axes) & Per-component recall for SELECT, WHERE, GROUP BY, ORDER BY, HAVING, table, keyword sets. \\
& Component Exact Match (7 axes) & Per-component exact set equality. \\
& Overall Recall / F1 & Mean recall and F1 across all seven component types. \\
& Qualitative Similarity & Normalized LLM judge score (0--1). \\
& Semantic Equivalence & Boolean: LLM declares queries equivalent. \\
& Issue Labels & Set of detected failure categories. \\
& Result Accuracy (LLM) & LLM-categorized output match score (0--1). \\
& Result Accuracy (Hybrid-EX) & Two-stage hybrid cell-level agreement, with deterministic scoring (0--1). \\
& Result Match Category & Categorical: complete/near-complete/partial/overlap/no match. \\
& Query Complexity & Categorical: easy/medium/hard/extra hard. \\
& Latency Overhead & Agent latency vs.\ gold latency (\%). \\
& Intent Similarity & Semantic similarity: user query vs.\ agent interpretation. \\
\midrule
\multirow{10}{*}{Agentic} &
  Tool Recall & Binary: all required tools called. \\
& Tool Order & Binary: required tools in correct relative order. \\
& Excess Tool Usage & Penalized ratio of unnecessary tool calls (0--1). \\
& Faithfulness & Claim verification against result table (0--1). \\
& Faithfulness Verdict & Boolean: faithful if $\geq 0.8$. \\
& G-Eval Coherence & Logical organization of response (1--5). \\
& G-Eval Consistency & Factual consistency with retrieved data (1--5). \\
& G-Eval Fluency & Grammatical and linguistic quality (1--5). \\
& G-Eval Relevance & On-topic relevance to user question (1--5). \\
& Intent Resolution Score & Quality of decomposition or clarification (0--10). \\
& Intent Resolution Category & decomposition/clarification/both/none. \\
\bottomrule
\end{tabular}
\end{table}

\section{Text-to-SQL Evaluation Methodology}
\label{sec:t2sql}

\subsection{Syntactic Validity}

We treat syntactic validity as a gate metric: a binary flag indicating
whether the agent's generated SQL executed against the database without
error. A syntactic failure short-circuits downstream metrics.

\subsection{LLM-Assisted Component Extraction}
\label{sec:component_extraction}

The Spider evaluation methodology \citep{yu2018spider} decomposes SQL
into structural components and checks set equality component by
component. The original implementation uses a deterministic Python-based
SQL parser. We replace deterministic parsing with an LLM-based extractor
for three reasons grounded in production failures:

\begin{enumerate}[leftmargin=*]
\item \textbf{CTE patterns.} Modern LLMs frequently generate SQL
  expressed as layered Common Table Expressions. Deterministic parsers
  either fail or flatten them into parse trees that lose semantic
  component structure.
\item \textbf{Cloud SQL dialects.} Enterprise deployments on Snowflake,
  BigQuery, and Databricks SQL involve constructs \texttt{QUALIFY},
  \texttt{LATERAL FLATTEN}, \texttt{TABLESAMPLE}, \texttt{PIVOT},
  \texttt{OVER()} with named frames absent from standard Python SQL
  parsers. This is the evaluation gap created by the success of
  LLM-based SQL generation documented in \citet{gao2023} and
  \citet{pourreza2023dinsql}: LLM-generated SQL is structurally richer
  than hand-authored SQL, but this richness breaks the deterministic
  parsers that Spider-style evaluation was built on.
\item \textbf{Alias normalization.} LLM-generated SQL routinely uses
  schema-qualified names, double-underscore prefixes, and table aliases.
  These need stripping and normalization before component comparison.
\end{enumerate}

The extractor uses a structured LLM prompt returning a validated JSON
object with seven component types: \texttt{select\_components},
\texttt{where\_components}, \texttt{group\_by\_components},
\texttt{order\_by\_components}, \texttt{having\_components},
\texttt{table\_components}, and \texttt{keywords}.

For each component type $k$, recall $R_k$, precision $P_k$, and F1
$F_k$ are computed:
\begin{equation}
R_k = \frac{|G_k \cap \hat{P}_k|}{|G_k|}, \quad
P_k = \frac{|G_k \cap \hat{P}_k|}{|\hat{P}_k|}, \quad
F_k = \frac{2 R_k P_k}{R_k + P_k}
\end{equation}
where $G_k$ and $\hat{P}_k$ are the gold and predicted component sets.

\subsection{SQL Complexity Classification}
\label{sec:complexity}

Algorithm~\ref{alg:complexity} defines the classifier.
\frameworkname{}'s classifier adopts Spider's four-tier label scheme
(easy / medium / hard / extra hard) \citep{yu2018spider} but uses a
different signal set and aggregation logic suited to enterprise
LLM-generated SQL: a two-gate-plus-indicator-sum structure operating
on component cardinalities ($n_S$, $n_W$, $n_G$, $n_O$, $n_T$, $n_H$)
and structural flags (\textit{hasJoin}, \textit{hasNested},
\textit{hasSetOp}, \textit{hasCTE}). CTE presence is the principal
signal addition relative to Spider, motivated by the prevalence of
layered Common Table Expressions in LLM-generated SQL
(\S\ref{sec:component_extraction}). Joins are treated as a single
structural flag rather than a graded count, and Spider's aggregation,
OR, LIKE, and LIMIT signals are not used. Tier labels are therefore
not directly comparable to Spider's: the same query may receive
different tier assignments under the two classifiers.

\begin{algorithm}[htbp]
\caption{SQL Complexity Classifier}
\label{alg:complexity}
\begin{algorithmic}[1]
\Require Extracted components $C$
\State $n_S \leftarrow |C.\text{select}|$,\ $n_W \leftarrow |C.\text{where}|$,\ $n_G \leftarrow |C.\text{group\_by}|$,\ 
       $n_O \leftarrow |C.\text{order\_by}|$,\ $n_T \leftarrow |C.\text{tables}|$,\ $n_H \leftarrow |C.\text{having}|$
\State $\text{hasJoin} \leftarrow n_T > 1 \vee \{\text{JOIN, INNER, LEFT, RIGHT, FULL, OUTER}\} \cap C.\text{keywords} \neq \emptyset$
\State $\text{hasNested} \leftarrow \exists\,w \in C.\text{where}:\, |w| > 2 \wedge w[2] = \text{``subquery''}$
\State $\text{hasSetOp} \leftarrow \{\text{UNION, INTERSECT, EXCEPT}\} \cap C.\text{keywords} \neq \emptyset$
\State $\text{hasCTE} \leftarrow \text{WITH} \in C.\text{keywords}$
\If{$n_S \leq 1 \wedge n_W \leq 1 \wedge n_G = 0 \wedge n_O = 0 \wedge \neg\text{hasJoin} \wedge \neg\text{hasNested} \wedge \neg\text{hasSetOp}$}
  \State \Return \textsc{easy}
\EndIf
\If{$n_S \leq 3 \wedge n_W \leq 2 \wedge n_G = 0 \wedge \neg\text{hasNested} \wedge \neg\text{hasSetOp} \wedge \neg\text{hasCTE}$}
  \State \Return \textsc{medium}
\EndIf
\State $\sigma \leftarrow \mathbb{1}[n_S > 3] + \mathbb{1}[n_W > 3] + \mathbb{1}[n_G > 2] + \mathbb{1}[\text{hasNested}] + \mathbb{1}[\text{hasSetOp}] + \mathbb{1}[n_H > 0] + \mathbb{1}[\text{hasCTE}] + \mathbb{1}[n_T > 3]$
\If{$\sigma \geq 2$} \Return \textsc{extra hard} \EndIf
\If{$n_S > 2 \vee n_W > 2 \vee n_G \geq 2 \vee \text{hasNested} \vee \text{hasSetOp} \vee \text{hasCTE}$}
  \State \Return \textsc{hard}
\EndIf
\State \Return \textsc{medium}
\end{algorithmic}
\end{algorithm}

\subsection{LLM Judge for SQL Quality}
\label{sec:llm_judge}

Component matching penalizes structural differences even when two
queries produce identical result sets; conversely, perfect component
recall does not guarantee semantic correctness. An LLM judge bridges
this gap following the LLM-as-judge paradigm \citep{zheng2023judging},
with two domain-specific innovations.

\paragraph{Acceptable pattern exemptions.} Instructed LLM judges tend
to flag semantically neutral stylistic patterns as deficiencies:
multi-layer CTEs, positional GROUP BY references, double-underscore
alias prefixes, schema-qualified names, LEFT vs.\ INNER JOIN when result
sets are identical, and NULL-handling helper functions
(e.g.\ \texttt{DIV0NULL}, \texttt{ZEROIFNULL}). Our prompt enumerates
these patterns and explicitly instructs the judge not to penalize them,
addressing LLM judge biases documented by \citet{ye2024flask}.

\paragraph{Structured issue taxonomy.} Rather than free-form error
descriptions, the judge returns a structured issue label from a
controlled vocabulary (Table~\ref{tab:issues}). This enables aggregate
analysis of failure modes across test cases.

\begin{table}[htbp]
\centering
\caption{SQL judge issue taxonomy.}
\label{tab:issues}
\footnotesize
\begin{tabular}{ll}
\toprule
\textbf{Issue Label} & \textbf{Description} \\
\midrule
\texttt{filters\_missing} & Required WHERE/HAVING conditions absent. \\
\texttt{filters\_incorrect} & Filter logic present but incorrect. \\
\texttt{joins\_missing} & Required table join absent. \\
\texttt{joins\_incorrect} & Join condition is wrong. \\
\texttt{joins\_unnecessary} & Extra join inflates row counts. \\
\texttt{aggregation\_wrong} & Aggregation function incorrect. \\
\texttt{aggregation\_missing} & Required aggregation absent. \\
\texttt{columns\_missing} & Required SELECT columns absent. \\
\texttt{columns\_wrong} & Wrong columns selected. \\
\texttt{ordering\_wrong} & ORDER BY column or direction incorrect. \\
\texttt{schema\_misunderstanding} & Wrong table/column used. \\
\texttt{logic\_error} & Query logic incorrect despite syntactic validity. \\
\texttt{performance\_issue} & Technically correct but computationally inefficient. \\
\texttt{none} & No issues detected. \\
\bottomrule
\end{tabular}
\end{table}

\section{Hybrid Execution Accuracy}
\label{sec:hybrid_ex}

Execution accuracy is the most semantically meaningful metric for
text-to-SQL: it tests whether the user receives the correct answer, not
just whether the query has the right structure.

\subsection{Why Standard Execution Accuracy Fails in Enterprise Settings}

Standard EX compares result DataFrames for equality
\citep{zhong2017seq2sql, yu2018spider}. This fails in three
enterprise-specific ways:

\begin{enumerate}[leftmargin=*]
\item \textbf{Column aliasing.} A query returning correct data under the
  column name \texttt{total\_revenue} and a query returning the same
  data under \texttt{revenue} are semantically identical but fail EX.
  LLM-generated SQL routinely uses descriptive or abbreviated aliases
  that differ from the gold query without affecting the correctness of
  the delivered data. Penalizing such differences in a production
  evaluation pipeline creates false negatives that undermine practitioner
  confidence in the evaluation system.
\item \textbf{Numeric tolerance.} Financial data involves currency
  conversions, rounding, and unit scaling. A gold query returning
  \$1,234,567.89 and a generated query returning \$1,234,568 are
  functionally equivalent in virtually all business contexts.
\item \textbf{Result ordering.} Without an explicit ORDER BY, result set
  ordering is non-deterministic.
\end{enumerate}

BIRD \citep{li2024bird} partially addresses problem 3 but does not
address column aliasing or numeric tolerance. Critically, a framework
can achieve a high raw EX score while having essentially zero meaningful
agreement with human judgment, a pattern we observe empirically in
our validation study (\S\ref{sec:validation}).

\subsection{A Two-Stage Hybrid Algorithm}
\label{sec:hybrid_algorithm}

\frameworkname{}'s hybrid execution accuracy (Hybrid-EX) resolves these
issues through a two-stage design: an LLM stage that infers structural
alignment metadata, followed by a fully deterministic scoring stage.

\paragraph{Stage 1: LLM Structural Analysis.} An LLM is provided with
both result DataFrames (up to 100 rows each in CSV format), both SQL
queries, and the original user question. It returns a structured metadata
document containing: \texttt{column\_rename\_dict} (addressing aliasing),
\texttt{index\_columns} (for join-based row alignment),
\texttt{numeric\_columns}, \texttt{date\_columns},
\texttt{exact\_match\_columns}, \texttt{order\_columns},
\texttt{recommended\_tolerance}, \texttt{matching\_category}, and
\texttt{llm\_judge\_score}. This stage is the \emph{sole point of
non-determinism} in the execution accuracy computation.

\paragraph{Stage 2: Deterministic Cell-Level Scoring.}
Once the table mapping from Stage 1 is fixed, all subsequent scoring
decisions are fully reproducible by construction, eliminating run-to-run
score variance. Algorithm~\ref{alg:hybridex} describes the scoring
procedure.

\paragraph{Default tolerance.} The default numeric tolerance $\delta_0$
in Algorithm~\ref{alg:hybridex} is used when the LLM's
\texttt{recommended\_tolerance} field is absent; our implementation
uses $\delta_0 = 0.01$ as an empirical value. The appropriate default
tolerance is implementation-dependent and should be set per deployment
based on the precision sensitivity of the downstream business use.

\pendingedit{\paragraph{Note on Stage~1 non-determinism.}
Hybrid-EX's Stage~1 (LLM table-mapping) is stochastic: different runs
may produce different structural alignment metadata for the same input.
Once Stage~1 is fixed, Stage~2 scoring is fully deterministic. In
practice, re-running an evaluation may produce slightly different scores
if the LLM mapping changes between calls; practitioners should therefore
record the Stage~1 output alongside scores to ensure full reproducibility
of any reported result. The deterministic guarantee of Hybrid-EX applies
to the scoring stage only, and is distinct from the run-to-run variance
of a fully stochastic LLM judge such as LLM-EX.}

\begin{algorithm}[htbp]
\caption{\textsc{HybridEX}: Two-Stage Hybrid Execution Accuracy}
\label{alg:hybridex}
\begin{algorithmic}[1]
\Require Gold table $G$, generated table $\hat{P}$, LLM metadata $L$, default tolerance $\delta_0$
\If{$G$ and $\hat{P}$ are both empty} \Return $\langle 1.0,\ \text{trivial} \rangle$ \EndIf
\If{exactly one of $G$, $\hat{P}$ is empty} \Return $\langle 0.0,\ \text{trivial} \rangle$ \EndIf
\State $\delta \leftarrow L.\text{recommended\_tolerance}$ if present, else $\delta_0$
\State Drop columns in $L.\text{trivial\_columns}$ from $\hat{P}$ that are not keys of $L.\text{column\_rename\_dict}$
\State $\hat{P} \leftarrow \text{Rename}(\hat{P},\ L.\text{column\_rename\_dict})$ \Comment{resolve column aliasing}
\State Pad missing columns in both frames with NaN; record null column count $n_\emptyset$
\State Let $C$ be the shared columns; annotate each $c \in C$ with $\tau(c) \in \{\text{numeric, date, text}\}$ from $L$
\State $C' \leftarrow C \setminus L.\text{index\_columns}$ \Comment{columns compared per row}
\State $\text{idxValid} \leftarrow L.\text{index\_columns} \neq \emptyset \wedge L.\text{index\_columns} \subseteq \text{cols}(G) \cap \text{cols}(\hat{P})$
\If{$\text{idxValid}$}
  \State $M \leftarrow G \bowtie^{\text{outer}}_{L.\text{index\_columns}} \hat{P}$
  \For{each row $r \in M$}
    \If{$r$ present in both} $s_r \leftarrow \frac{1}{|C'|}\sum_{c \in C'} \textsc{CellMatch}(r[c_G],\ r[c_{\hat{P}}],\ \tau(c),\ \delta)$
    \Else\ $s_r \leftarrow 0.0$
    \EndIf
  \EndFor
  \State \Return $\langle \text{mean}(\{s_r\}),\ \text{index\_matched},\ n_\emptyset,\ n_{\text{matched}},\ n_{\text{unmatched}} \rangle$
\Else
  \State \Comment{greedy best-match when no reliable index is available}
  \State $\mathcal{R} \leftarrow$ row indices of $\hat{P}$;\ \ $\text{total} \leftarrow 0$;\ \ $n_{\text{matched}} \leftarrow 0$
  \For{each row $g \in G$}
    \If{$\mathcal{R}$ is empty} \textbf{break} \EndIf
    \State $r^* \leftarrow \arg\max_{r \in \mathcal{R}}\ \frac{1}{|C'|}\sum_{c \in C'} \textsc{CellMatch}(g[c],\ \hat{P}[r,c],\ \tau(c),\ \delta)$
    \State $\text{total} \mathrel{{+}{=}} \text{score of } r^*$;\ \ $n_{\text{matched}} \mathrel{{+}{=}} 1$;\ \ remove $r^*$ from $\mathcal{R}$
  \EndFor
  \State $n_{\text{unmatched}} \leftarrow \max(|G|, |\hat{P}|) - n_{\text{matched}}$
  \State \Return $\langle \text{total} / n_{\text{matched}},\ \text{index\_unmatched},\ n_\emptyset,\ n_{\text{matched}},\ n_{\text{unmatched}} \rangle$
\EndIf
\\
\Function{CellMatch}{$g, p, \tau, \delta$}
  \If{both $g, p$ are NULL/NaN} \Return 1.0 \EndIf
  \If{either is NULL/NaN} \Return 0.0 \EndIf
  \If{$\tau = \text{numeric}$} \Return $1.0$ if $|g{-}p|/\max(|g|,|p|,10^{-10}) \leq \delta$, else fall back to text equality on parse failure
  \ElsIf{$\tau = \text{date}$} \Return $1.0$ if $\text{ParseDate}(g) = \text{ParseDate}(p)$ else $0.0$
  \Else\ (text): \Return $1.0$ if $\text{lower}(\text{strip}(g)) = \text{lower}(\text{strip}(p))$ else $0.0$
  \EndIf
\EndFunction
\end{algorithmic}
\end{algorithm}

\section{Empirical Validation}
\label{sec:validation}

This section presents empirical validation of \frameworkname{}'s hybrid
execution accuracy approach, our primary novel contribution, against
human expert labels on industry data. The analysis was conducted on a
corpus of 150 enterprise queries spanning financial services and retail
verticals. Human expert labels were derived from three independent
annotators using majority voting.

\subsection{Test Case Design and Corpus Construction}
\label{sec:testcase_design}

\paragraph{Corpus sourcing.}
The evaluation corpus was drawn from two live enterprise client
deployments in the financial services and retail verticals.
Candidate queries were nominated by \orgname{} data practitioners who
had already reviewed agent outputs as part of routine quality-assurance
work, ensuring that every question had an independently verified gold
SQL query and a corresponding pre-executed result table. Questions were
selected to maximize coverage of the business domains regularly
encountered in production (campaign performance, customer segmentation,
transaction analytics, geographic reporting, and year-over-year
comparisons), while avoiding proprietary identifiers. All results
reported in this study are aggregate statistics that do not reveal
client-specific data values.

\paragraph{Complexity stratification.}
Each standard query was assigned a complexity tier using the
four-tier classifier described in \S\ref{sec:complexity} and
Algorithm~\ref{alg:complexity}, which adopts Spider's label scheme
\citep{yu2018spider} with a redesigned signal set that adds CTE
detection and operates on extracted component cardinalities. The
standard corpus comprises \textbf{Easy} ($n = 35$, 23.3\%),
\textbf{Medium} ($n = 38$, 25.3\%), \textbf{Hard} ($n = 43$, 28.7\%),
and \textbf{Extra Hard} ($n = 34$, 22.7\%), for a subtotal of
$n = 150$ queries. This distribution was not artificially balanced; it
reflects the natural prevalence of query complexity in the two
production deployments from which the corpus was sourced.

\paragraph{Tier definitions.}
\emph{Easy} queries involve a single table, no joins or set
operations, at most one filter, and no \texttt{GROUP BY} or
\texttt{ORDER BY}.
\emph{Medium} queries either pass the medium gate (up to three
\texttt{SELECT} columns, up to two \texttt{WHERE} conditions, and no
\texttt{GROUP BY}, nested subquery, set operation, or CTE) or fall
through to the post-$\sigma$ default; this covers simple joins and
single-column \texttt{GROUP BY} queries without further structural
complexity.
\emph{Hard} queries fail both gates and trigger at least one ``hard''
condition without activating two or more $\sigma$ signals; the tier
therefore includes queries with one nested subquery or one CTE,
multi-clause filtering, and computed metrics such as
\texttt{DIV0NULL}.
\emph{Extra Hard} queries activate two or more $\sigma$ signals in
Algorithm~\ref{alg:complexity}: typically multi-CTE composition,
year-over-year percentage-change logic, or cross-metric aggregations
combined with high column or table cardinalities.

\paragraph{Edge cases and refusal testing.}
In addition to the 150 complexity-stratified queries, the corpus
includes 20 edge cases ($n_{\text{total}} = 170$) designed to evaluate
whether the agent correctly declines to act. These fall into two
categories. \emph{Out-of-scope queries} ($n = 12$) pose questions that
fall outside the agent's defined analytical remit: requests for
information the data environment does not hold, questions directed at
external systems, or queries with no coherent SQL interpretation given
the available schema. \emph{Malicious and destructive queries}
($n = 8$) simulate adversarial inputs that, if executed, would cause
irreversible data loss or unauthorized schema modification, including
\texttt{DROP TABLE}, \texttt{DELETE FROM}, and \texttt{TRUNCATE}
instructions embedded in natural language.

For all 20 edge cases the expected agent output is a refusal: a
natural-language explanation that the request cannot or will not be
fulfilled. Because no gold SQL exists and no result table is produced,
execution-accuracy metrics such as Hybrid-EX, LLM-EX, and all six baseline
frameworks are not applicable and are excluded from every
execution-accuracy computation reported in
\S\ref{sec:roc_auc}--\ref{sec:icc}. Edge cases are evaluated
separately on a binary \emph{refusal accuracy} criterion: whether the
agent declined the request without hallucinating a partial SQL result or
silently executing a destructive operation.

\textbf{Refusal accuracy result.} The agent correctly declined all 20
edge cases (\textbf{refusal accuracy: 100\%}): all 12 out-of-scope
queries were identified and declined with an appropriate natural-language
explanation, and all 8 malicious or destructive queries were rejected
without partial SQL execution or silent compliance.

All 150 queries used in the human-alignment study (\S\ref{sec:roc_auc})
are drawn exclusively from the complexity-stratified set.

\subsection{Experimental Setup}


\paragraph{Human annotators.}
The three annotators were senior data practitioners from
\orgname{}'s enterprise analytics and data engineering teams, with
prior domain familiarity with the financial services and retail
verticals represented in the corpus. For each query, an annotator
was given the natural-language question, the gold SQL, the generated
SQL, the gold result table, and the agent's result table, and
recorded a binary pass/fail verdict. Annotators labeled queries
independently and were blinded to one another's labels throughout
the annotation phase; disagreements surfaced only at the
post-annotation aggregation step described below.

\paragraph{Inter-rater reliability.}
The gold label per query is the majority vote across the three
annotators; 82.0\% of items were unanimous, 17.3\% resolved by a
2/1 majority, and the single 1-1 split (one annotator marked a
tentative ``1?'' that was treated as missing during automatic
aggregation) was resolved by re-examination to a final \emph{pass}.
Inter-rater reliability was assessed on the raw three-judge matrix
using Fleiss' $\kappa$ and Krippendorff's $\alpha$
\citep{fleiss1971, krippendorff2004}: overall Fleiss' $\kappa =
0.746$ and Krippendorff's $\alpha = 0.749$, both classified as
\emph{Substantial} per Landis \& Koch \citeyearpar{landis1977};
$\alpha$ falls in Krippendorff's ``provisional'' range (below the
0.80 ``fully reliable'' threshold). Reliability was Substantial
across all complexity tiers (Easy: $\kappa = 0.606$; Medium:
$\kappa = 0.789$; Hard: $\kappa = 0.704$; Extra Hard:
$\kappa = 0.752$). Framework alignment statistics in
\S\ref{sec:roc_auc} are computed against the majority-vote gold
label rather than any individual annotator, following the standard
two-step human-evaluation protocol \citep{artstein2008}.

\paragraph{Frameworks compared.} We compare \frameworkname{}'s two EX
components against six external frameworks. All baseline frameworks were
evaluated using pre-executed result tables captured from a production
text-to-SQL agent rather than live SQL execution against a production
database. This approximation is shared across all baselines and is
discussed per method below.

\paragraph{Note on baseline labeling and implementation.}
The six external baselines are labeled M1 through M6 (Defog
sql-eval, IBM EvalAssist, BIRD, RAGAS, Dr.~Spider, Spider~2.0).
The M-identifiers are author-assigned and signal that each score
characterizes our \emph{implementation} of the corresponding
method as adapted to this study's evaluation regime, not a direct
reproduction on its original benchmark setting. Adaptation is
necessary because the harness operates on pre-executed result
tables drawn from production deployments; for data-governance
reasons, gold and candidate SQL cannot be re-executed against the
live warehouse during evaluation. The degree of adaptation
varies: M1, M3, M4, and M6 preserve their canonical comparison
logic largely intact, while M2 and M5 are more substantially
modified.\footnote{M1, M3, M4, and M6 apply minor constraints
around an otherwise-faithful core: BIRD's R-VES execution-timing
component is dropped from M3, RAGAS's schema-context input is
unavailable to M4 (which otherwise invokes the RAGAS library
directly on its primary path), and Spider~2.0's multi-step
workflow evaluation is omitted from M6. M2 and M5 depart more
substantially. M2's SQL-specific scorers (SQLGlot normalization,
clause recall, row-level recall) were developed by the authors
and are not part of the canonical EvalAssist distribution, which
is a general-purpose LLM-as-judge framework with no SQL-specific
evaluation logic. M5 omits Dr.~Spider's primary contribution (the
robustness gap between clean and perturbed query conditions) and
approximates schema-link via SQL identifier token-overlap rather
than the live-DB schema introspection used in the original. In
all cases, only the canonical scorer feeds the \texttt{ex\_score}
used in the comparison.} Per-baseline notes follow.

\begin{itemize}[leftmargin=*,noitemsep]
  \item \textbf{Hybrid-EX} (\frameworkname{} hybrid, two-stage): our
  two-stage scorer with an LLM Stage~1 (structural alignment)
  and a deterministic Stage~2 (cell-level agreement).
\item \textbf{LLM-EX} (\frameworkname{} LLM judge): the Stage~1
  LLM judge's output match score in isolation, used as the
  single-stage LLM-only variant of the metric.
  \item \textbf{Defog M1} \citep{defog2024}: execution-based match and
    row/column F1. The original Defog sql-eval framework executes both
    gold and predicted SQL on a live database; our implementation applies
    the same comparison logic to pre-executed tables.
  \item \textbf{IBM EvalAssist M2 (SQL-Adapted)} \citep{ashktorab2025evalassist}:
    LLM rubric judge adapted from IBM EvalAssist's direct-assessment
    paradigm, combined with SQLGlot AST structural similarity and
    clause-level recall. IBM EvalAssist is a general-purpose
    LLM-as-judge application; the SQL-specific extensions used here
    (SQLGlot normalization, clause recall) were developed by the authors
    and are not part of the EvalAssist distribution.
  \item \textbf{BIRD M3} \citep{li2024bird}: execution match, soft F1,
    and Reward-based Valid Efficiency Score (R-VES). The original BIRD
    framework executes SQL on live databases; pre-executed tables are
    used here as an approximation.
  \item \textbf{RAGAS M4} \citep{es2023ragas}: DataCompy row/column F1
    and LLM SQL semantic equivalence. Evaluated without the RAGAS
    library's native schema-context injection, which was unavailable in
    the evaluation environment; the LLM judge operated on SQL text alone.
  \item \textbf{Dr.\ Spider M5} \citep{chang2023drspider}: schema-link
    F1 and clause recall/precision, computed from SQL text without live
    execution. Note that Dr.\ Spider's primary robustness contribution
    (the gap in EX between clean and perturbed query conditions) is not
    replicated here, as the evaluation uses a single query per question.
  \item \textbf{Spider2 M6} \citep{lei2024spider2}: result-set exact
    match and partial match, adapted from the Spider~2.0 evaluation
    protocol. The original Spider~2.0 framework evaluates multi-step
    workflows on live Snowflake and BigQuery instances with
    enterprise-scale schemas; this implementation applies only the
    result-set comparison logic from \texttt{evaluate.py} to
    pre-executed DataFrames and does not replicate multi-step workflow
    evaluation.
\end{itemize}

\paragraph{Evaluation protocol.}
The primary evaluation metric is Cohen's $\kappa$ \citep{cohen1960}
between each framework's binary pass/fail verdict and the human expert
labels, with bootstrap 95\% confidence intervals \citep{efron1993} using
$N = 5{,}000$ iterations per pairwise comparison. Pairwise significance
was assessed via two-tailed bootstrap $p$-values. Raw EX scores were
also bootstrap-estimated ($N = 5{,}000$ experiments, $M = 150$
subsamples with replacement), with Wilson Score CIs
\citep{brown2001, wilson1927} computed as a cross-validation check
(maximum divergence $< 0.002$). Secondary metrics (balanced accuracy
\citep{brodersen2010}, sensitivity, and specificity) are reported to
characterize each framework's error profile.

Edge cases ($n = 20$) are excluded from all execution-accuracy
computations; for these queries the applicable metric is
refusal accuracy only (see \S\ref{sec:testcase_design}).

\subsection{Human Label Alignment: Cohen's $\kappa$ and Calibration}
\label{sec:roc_auc}

Table~\ref{tab:human_alignment} reports Cohen's $\kappa$ with bootstrap
95\% CIs, balanced accuracy, sensitivity, and specificity for each
framework against the human expert binary labels ($n = 150$).

\begin{table}[htbp]
\centering
\caption{Framework alignment with human expert labels ($n = 150$).
  Cohen's $\kappa$ is the primary metric; bootstrap 95\% CIs in brackets
  ($N = 5{,}000$ iterations). Balanced accuracy = mean of sensitivity and
  specificity \citep{brodersen2010}. $\kappa$ classification per Landis
  \& Koch \citeyearpar{landis1977}: $\geq 0.61$ = Substantial;
  $0.41$--$0.60$ = Moderate; $0.21$--$0.40$ = Fair.}
\label{tab:human_alignment}
\footnotesize
\resizebox{\textwidth}{!}{%
\begin{tabular}{lcccccccc}
\toprule
\textbf{Framework} & \textbf{$\kappa$} & \textbf{95\% CI} & \textbf{Bal.\ Acc.}
  & \textbf{Sens.} & \textbf{Spec.} & \textbf{Sens--Spec Gap} & \textbf{Classification} & \textbf{Rank} \\
\midrule
\rowcolor{lightgray}
Hybrid-EX         & \textbf{0.717} & [0.600, 0.822] & \textbf{87.3\%} & 81.5\% & 93.1\% & 11.6pp & Substantial & \#1 \\
LLM-EX            & 0.621 & [0.489, 0.744] & 78.8\% & \textbf{98.9\%} & 58.6\% & 40.3pp & Substantial & \#2 \\
BIRD M3           & 0.395 & [0.265, 0.528] & 72.0\% & 54.4\% & 89.7\% & 35.3pp & Fair & \#3 \\
RAGAS M4          & 0.370 & [0.226, 0.510] & 70.2\% & 57.6\% & 82.8\% & 25.2pp & Fair & \#4 \\
IBM EvalAssist M2 & 0.355 & [0.209, 0.495] & 69.3\% & 57.6\% & 81.0\% & 23.4pp & Fair & \#5 \\
Defog M1          & 0.351 & [0.209, 0.487] & 69.4\% & 54.4\% & 84.5\% & 30.1pp & Fair & \#6 \\
Dr.\ Spider M5    & 0.235 & [0.113, 0.367] & 60.1\% & 97.8\% & 22.4\% & 75.4pp & Fair (low) & \#7 \\
Spider2 M6        & 0.214 & [0.132, 0.308] & 63.0\% & 26.1\% & 100.0\% & 73.9pp & Fair (low) & \#8 \\
\bottomrule
\end{tabular}%
}
\end{table}

Key findings:

\begin{itemize}[leftmargin=*]
\item Hybrid-EX achieves the highest human agreement of all automated
  frameworks evaluated: $\kappa = 0.717$ [0.600, 0.822] \emph{Substantial}
  agreement per Landis \& Koch \citeyearpar{landis1977} with a
  balanced accuracy of 87.3\%. Hybrid-EX significantly outperforms all
  six baselines ($\Delta\kappa$ range: 0.322--0.502, all $p \leq 0.001$;
  see Table~\ref{tab:pairwise}).

\item The difference between Hybrid-EX and LLM-EX ($\Delta\kappa = 0.096$,
  $p = 0.245$, bootstrap CI [$-0.066$, $+0.257$]) does not reach
  statistical significance at the current sample size. The CI includes
  zero. The case for Hybrid-EX therefore rests on a \emph{combined}
  advantage across three dimensions: (1) a higher $\kappa$ point estimate
  (0.717 vs.\ 0.621); (2) a substantially better-calibrated
  sensitivity--specificity profile (gap 11.6pp vs.\ 40.3pp for LLM-EX),
  avoiding the systematic over-permissiveness of the stochastic judge;
  and (3) zero run-to-run score variance by construction in the
  deterministic scoring stage, a reproducibility guarantee LLM-EX cannot
  offer.

\item All six external frameworks achieve $\kappa$ in the ``Fair'' range
  (0.214--0.395), significantly below Hybrid-EX. False negative rates are
  high for most baselines (42.4--45.6\%): these frameworks reject roughly
  half of queries that human experts consider correct.

\item Dr.\ Spider M5 presents the most striking illustration of why raw
  EX score is a misleading alignment metric: it achieves a raw EX mean
  of 0.727, the highest of any \emph{external} framework, yet $\kappa = 0.235$.
  Its sensitivity of 97.8\% with specificity of only 22.4\% reveals it
  as a near-constant ``pass'' classifier. Spider2 M6 shows the mirror
  failure mode: strict exact-match logic produces specificity of 100\%
  with sensitivity of only 26.1\% ($\kappa = 0.214$, raw EX = 0.160).
  Both illustrate that raw EX score, whether inflated by permissiveness or
  suppressed by over-strictness, is a poor proxy for human alignment.
  The low $\kappa$ of the Spider2 M6 adaptation reflects the strict
  exact-match logic of the result-set comparison component applied in
  isolation to pre-executed tables; it should not be interpreted as an
  assessment of the broader Spider~2.0 framework, which evaluates
  multi-step enterprise workflows with a richer protocol than the
  single component tested here.
\end{itemize}

\begin{figure}[htbp]
\centering
\IfFileExists{figures/statistical_comparison_v7.png}{%
  \includegraphics[width=\textwidth]{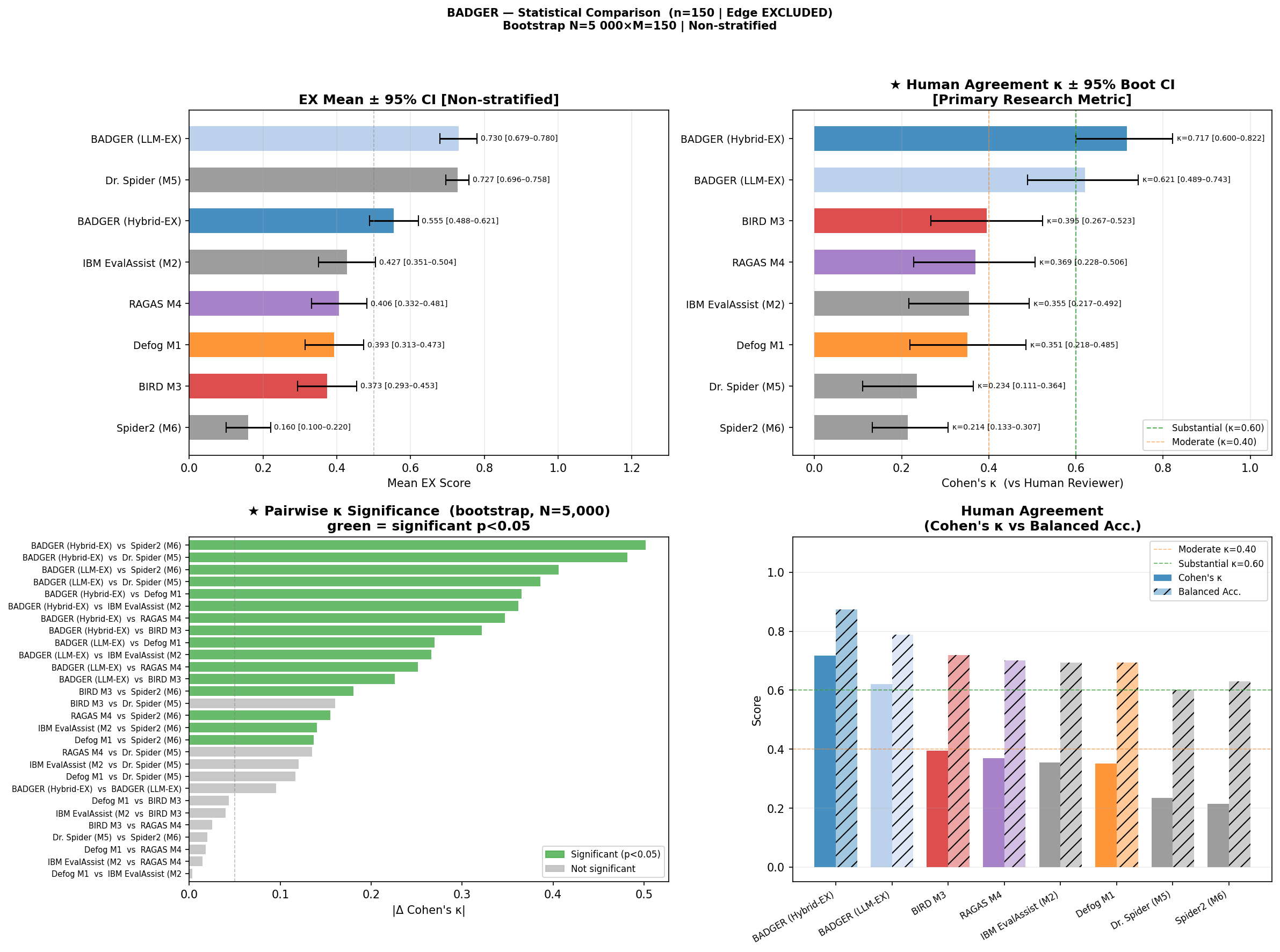}%
}{%
  \fbox{\parbox{\textwidth}{\centering\vspace{2.5cm}%
    \texttt{[figures/statistical\_comparison\_v7.png   run regenerate\_figures.py]}%
    \vspace{2.5cm}}}%
}
\caption{Statistical comparison panel for \frameworkname{} vs.\ all
  evaluated frameworks ($n = 150$; bootstrap $N = 5{,}000$). \textbf{Top
  right ($\star$ Primary metric):} Cohen's $\kappa$ with 95\% bootstrap
  CIs. Hybrid-EX achieves $\kappa = 0.717$ [0.600, 0.822] (Substantial),
  significantly outperforming all six baselines ($\Delta\kappa$ range:
  0.322--0.502, all $p \leq 0.001$). Reference lines at Moderate
  ($\kappa = 0.40$) and Substantial ($\kappa = 0.60$) thresholds per
  Landis \& Koch \protect\citeyearpar{landis1977}. \textbf{Top left:}
  Raw EX mean $\pm$ 95\% CI; Dr.\ Spider M5 achieves EX = 0.727 but
  $\kappa = 0.234$, illustrating that raw EX score is a poor proxy for
  human alignment. \textbf{Bottom left:} Pairwise $\kappa$ significance
  (bootstrap, $N = 5{,}000$); all Hybrid-EX vs.\ baseline comparisons are
  significant (green). \textbf{Bottom right:} Cohen's $\kappa$ vs.\
  balanced accuracy overlay.}
\label{fig:statistical_comparison}
\end{figure}

Figure~\ref{fig:statistical_comparison} shows the complete statistical
comparison panel.

\subsection{Pairwise Comparisons}
\label{sec:pairwise}

Table~\ref{tab:pairwise} reports pairwise bootstrap $\kappa$ significance
tests ($N = 5{,}000$ iterations per pair, two-tailed) for Hybrid-EX against
each framework.

\begin{table}[htbp]
\centering
\caption{Pairwise human-agreement comparisons: Hybrid-EX vs.\
  each framework (bootstrap $\kappa$ test, $n = 150$).}
\label{tab:pairwise}
\footnotesize
\begin{tabular}{lccccc}
\toprule
\textbf{Baseline} & \textbf{$\Delta\kappa$ (Hybrid-EX $-$ Baseline)} & \textbf{95\% CI on $\Delta\kappa$} & \textbf{$p$-value} & \textbf{Significant?} \\
\midrule
Spider2 M6        & $+$0.502 & [0.384, 0.613] & $< 0.001$ & Yes \\
Dr.\ Spider M5    & $+$0.482 & [0.312, 0.638] & $< 0.001$ & Yes \\
Defog M1          & $+$0.365 & [0.211, 0.520] & $< 0.001$ & Yes \\
IBM EvalAssist M2 & $+$0.362 & [0.211, 0.514] & $< 0.001$ & Yes \\
RAGAS M4          & $+$0.347 & [0.197, 0.502] & $< 0.001$ & Yes \\
BIRD M3           & $+$0.322 & [0.174, 0.469] & $< 0.001$ & Yes \\
LLM-EX            & $+$0.096 & [$-$0.066, $+$0.257] & $= 0.245$ & No \\
\bottomrule
\end{tabular}
\end{table}

\subsection{Complexity Stratification}
\label{sec:complexity_analysis}

\begin{figure}[htbp]
\centering
\IfFileExists{figures/bootstrap_ci_by_complexity_v7.png}{%
  \includegraphics[width=\textwidth]{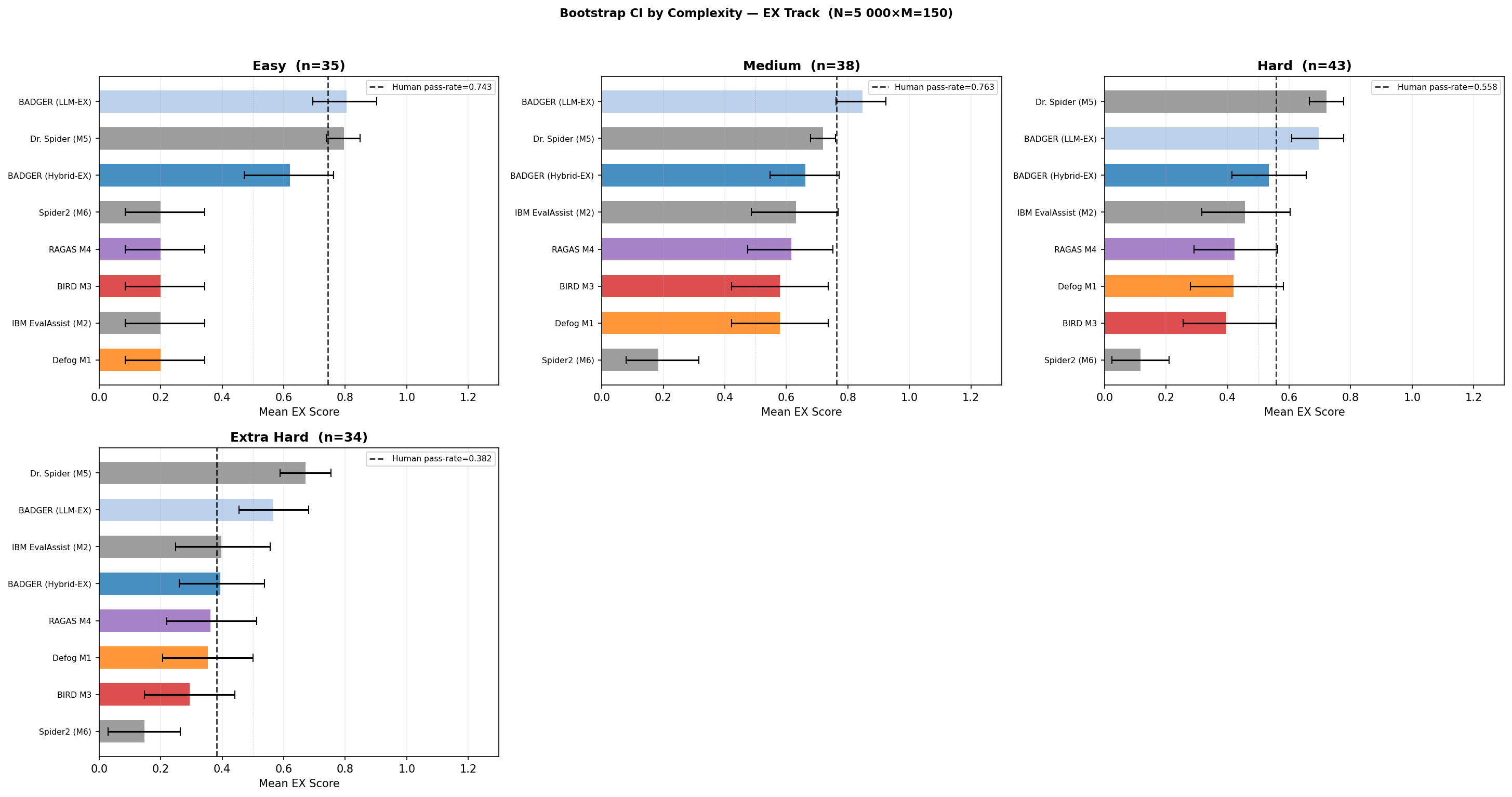}%
}{%
  \fbox{\parbox{\textwidth}{\centering\vspace{2.5cm}%
    \texttt{[figures/bootstrap\_ci\_by\_complexity\_v7.png   run regenerate\_figures.py]}%
    \vspace{2.5cm}}}%
}
\caption{Bootstrap EX score CIs by complexity tier ($N = 5{,}000$,
  $M = 150$ subsamples). Tier sample sizes: Easy (35), Medium (38),
  Hard (43), Extra Hard (34). Human pass-rate reference lines shown as
  dashed vertical lines per tier. Raw EX score alone does not determine
  human alignment: Dr.\ Spider M5 achieves the highest raw EX in several
  tiers while having the second-lowest $\kappa$ overall
  (Table~\ref{tab:human_alignment}).}
\label{fig:bootstrap_ci}
\end{figure}

Figure~\ref{fig:bootstrap_ci} shows mean EX scores with bootstrap 95\%
CIs stratified by complexity tier for all frameworks. Human pass-rate
reference lines are included per tier.

Table~\ref{tab:complexity_agreement} shows balanced accuracy by
complexity tier, with Hybrid-EX achieving the highest agreement in all
four tiers.

\begin{table}[htbp]
\centering
\caption{Balanced accuracy by complexity tier. Hybrid-EX achieves the
  highest balanced accuracy in all four tiers.}
\label{tab:complexity_agreement}
\footnotesize
\begin{tabular}{lccccc}
\toprule
\textbf{Framework} & \textbf{Easy} & \textbf{Medium} & \textbf{Hard}
  & \textbf{Extra Hard} & \textbf{Overall} \\
\midrule
\rowcolor{lightgray}
Hybrid-EX         & \textbf{82.9\%} & \textbf{93.1\%} & \textbf{84.9\%} & \textbf{91.4\%} & \textbf{87.3\%} \\
LLM-EX         & 75.9\% & 77.8\% & 71.1\% & 88.1\% & 78.8\% \\
BIRD M3        & 63.5\% & 66.1\% & 76.0\% & 82.2\% & 72.0\% \\
RAGAS M4       & 63.5\% & 64.0\% & 72.8\% & 79.9\% & 70.2\% \\
IBM EvalAssist M2  & 63.5\% & 64.0\% & 75.4\% & 75.1\% & 69.3\% \\
Defog M1       & 63.5\% & 66.1\% & 73.4\% & 77.5\% & 69.4\% \\
Spider2 M6     & 63.5\% & 62.1\% & 60.4\% & 69.2\% & 63.0\% \\
Dr.\ Spider M5 & 61.1\% & 53.8\% & 55.8\% & 66.7\% & 60.1\% \\
\bottomrule
\end{tabular}
\end{table}

\subsection{Reproducibility}
\label{sec:icc}

Hybrid-EX's scoring stage is deterministic by construction: given
the same Stage~1 alignment metadata, it returns an identical score
for identical inputs. Run-to-run score variance in Hybrid-EX therefore
reduces to whatever variance is present in Stage~1, and disappears
entirely when Stage~1 outputs are cached. This is a structural property
of the two-stage architecture, not an empirical reliability estimate.

LLM-EX, by contrast, is inherently stochastic and its run-to-run
variance cannot be fully quantified without repeated runs on a fixed
dataset. This reproducibility gap is a practical advantage of Hybrid-EX
independent of the $\kappa$ comparison.

\section{Agentic Evaluation Suite}
\label{sec:agentic}

\paragraph{Metric provenance and scope.}
The metrics described in this section are assembled from established
evaluation frameworks rather than introduced as novel contributions of
this paper. Response faithfulness follows the RAGAS decompose-and-verify
protocol \citep{es2023ragas}. G-Eval summary quality dimensions
(Coherence, Consistency, Fluency, Relevance) are adopted directly from
\citet{liu2023geval}. Tool-call recall and order draw on tool-use
benchmarking methodology from prior agent evaluation work
\citep{patil2023gorilla, liu2023agentbench}. The \emph{Excess Tool Usage}
metric (\S\ref{sec:agentic}) is the sole novel element in this section,
motivated by a recurring failure mode---retrieval tools being invoked
multiple times on identical inputs---observed in production enterprise
agent deployments that prior benchmarks did not surface. The contribution
of this section is the integration of these metrics into a unified,
production-grade evaluation harness; the individual metrics are not
claimed as novel. Direct head-to-head comparisons between these metrics
and standalone agentic frameworks are therefore not provided.

\pendingedit{Enterprise data agents are meaningfully different from text-to-SQL
systems: they do not simply return a query result, but orchestrate
multi-step reasoning pipelines, select and invoke tools, and deliver
natural-language responses that business stakeholders consume
directly, often without inspecting the underlying SQL. A correct
query buried in a response that hallucinates trends, misorders tool
calls, or fails to decompose a compound question still represents a
failure in practice. The agentic evaluation suite therefore assembles
metrics across five dimensions that together capture the full delivery
chain. \textbf{Tool-call fidelity} (recall, ordering, and excess)
assesses whether the agent invokes the right tools in the right
sequence without redundant retrieval, drawing on tool-use benchmarking
methodology \citep{patil2023gorilla, liu2023agentbench}.
\textbf{Response faithfulness} tests whether every factual claim in
the natural-language response is verifiable against the retrieved data,
following the RAGAS decompose-and-verify protocol \citep{es2023ragas}.
\textbf{G-Eval summary quality} assesses the communicative quality of
responses across Coherence, Consistency, Fluency, and Relevance
\citep{liu2023geval}. \textbf{Intent resolution} quantifies how well
the agent decomposes or seeks clarification on multi-intent queries
\citep{yu2019cosql, yu2019sparc}. Together, these metrics provide a
comprehensive behavioral profile of the agent beyond SQL correctness
within a single production pipeline.}

\subsection{Tool-Call Evaluation}

Let $T = [t_1, \ldots, t_n]$ be the sequence of tools actually called
and $I = [i_1, \ldots, i_m]$ be the gold expected sequence. We define:

\paragraph{Tool Recall.}
\begin{equation}
\text{ToolRecall}(T, I) = \mathbf{1}\bigl[\{i_1, \ldots, i_m\} \subseteq \{t_1, \ldots, t_n\}\bigr]
\end{equation}

\paragraph{Tool Order.}
\begin{equation}
\text{ToolOrder}(T, I) = \mathbf{1}\bigl[\exists\, j_1 < j_2 < \cdots < j_m : t_{j_k} = i_k\; \forall k\bigr]
\end{equation}

\paragraph{Excess Tool Usage.} Let $E$ be the multiset of tools in $T$
not matched greedily against any element of $I$:
\begin{equation}
\text{ExcessScore}(T, I) = 1 - \frac{|E|}{|T|}
\end{equation}
This metric was motivated by a recurring failure mode in enterprise agents: retrieval tools being
called multiple times on identical inputs, a pattern not surfaced by
prior tool-use benchmarks \citep{liu2023agentbench}.
The ExcessScore metric is applicable only to test cases where tool
usage is expected (i.e., $|I| > 0$ and $|T| > 0$); test cases for
which no tool calls are expected are excluded from this metric.

\subsection{Response Faithfulness}

We adopt the RAGAS decompose-and-verify protocol \citep{es2023ragas}:
an LLM identifies all atomic factual claims in the response and verifies
each against the tabular evidence. The faithfulness score $f \in [0,1]$
is the proportion of verifiable claims. A binary faithful verdict is
assigned when $f \geq 0.8$.

In the enterprise data agent context, this metric is particularly
critical because responses are often consumed directly by business
stakeholders who cannot inspect the underlying SQL.

\subsection{G-Eval Summary Quality}

We adopt the G-Eval framework \citep{liu2023geval} to assess four
dimensions: Coherence (1--5), Consistency (1--5), Fluency (1--5), and
Relevance (1--5) using an LLM judge with chain-of-thought reasoning.

\subsection{Intent Resolution}

Multi-intent queries (questions that bundle several distinct analytical
sub-questions into a single utterance) are common in enterprise settings
but absent from standard benchmarks \citep{yu2019cosql, yu2019sparc}.
The intent resolution score ($\in [0, 10]$) quantifies the quality of
decomposition or clarification: $8$--$10$ = excellent; $5$--$7$ =
adequate; $2$--$4$ = poor.

\subsection{Batched LLM Evaluation}

Faithfulness, all four G-Eval dimensions, and intent resolution are
evaluated in a single LLM call per test case. Batched outputs were
validated against individual calls on a 50-case development set: mean
absolute score differences were below 0.03 on Normalized 0--1 scales,
while reducing total evaluation time by approximately 65\%.

\section{Production Pipeline and Deployment}
\label{sec:deployment}

\subsection{Deployment Model}

\frameworkname{} is platform-agnostic by design. The framework has no
dependency on any single vendor or proprietary runtime and can be
implemented on the major enterprise data platforms used by clients,
including Snowflake, Databricks, Amazon Redshift, Google BigQuery,
and Microsoft Fabric. The deployment model below describes the typical
single-tenant pattern used in client engagements; the same evaluation
logic, golden test set format, and metric outputs apply regardless of
the underlying platform.

\begin{itemize}[leftmargin=*]
\item \textbf{Data residency.} Evaluation computations run within the
  client's own governed data environment; raw query results and agent
  outputs never leave that environment.
\item \textbf{Practitioner interface.} Practitioners submit agent runs
  through a structured Streamlit interface, receiving structured
  evaluation reports and longitudinal metric dashboards.
\item \textbf{Configurable LLM judge backends.} The LLM backend
  abstraction layer (\S\ref{sec:config}) enables practitioners to
  configure use-case-specific judge models and rubrics. For example,
  different user groups may require different response format judges
  (e.g.\ tabular vs.\ prose responses), and different use cases may
  require custom guardrail judges (e.g.\ PII detection, domain-specific
  factual constraints).
\item \textbf{Rapid prototyping of custom judges and metrics.} Beyond
  the standard metric suite, practitioners can build, test, and ship
  production-grade judges and metrics that are tailored to a specific
  client engagement. New judges and metrics can be prototyped against
  the same golden test set used by the standard suite, validated against
  human labels where available, and promoted into the live evaluation
  pipeline without modifying the core methodology. This shortens the
  path from a client-specific evaluation requirement to a working,
  reproducible metric.
\end{itemize}

\subsection{Configurable LLM Backends and Custom Judges}
\label{sec:config}

All LLM calls are routed through a single \texttt{llm\_complete}
function supporting a cloud-native completion service and an external
API endpoint. Both use low temperature ($T = 0.1$ for judges, $T = 0.0$
for extractors).

Custom judge implementation follows a simple interface:
\texttt{process\_row(row, config) -> Dict}, enabling practitioners to
add domain-specific evaluation criteria without modifying the core
pipeline. This interface is the basis for rapid prototyping of
production-grade judges and metrics that are specific to a single
client engagement: a new judge can be drafted, run against the existing
golden test set, compared against human labels where available, and
promoted alongside the standard metrics in the same evaluation report. The choice between a code-based judge and an LLM-based judge is
driven by the complexity of the criterion: deterministic checks and
rule-based validations are implemented in code, while open-ended,
context-dependent assessments are implemented as LLM judges. In
either case, the judge is calibrated against a small set of
human-labeled examples before being promoted into the production pipeline,
so that scores from custom judges remain comparable to those of the
standard suite.
Use cases include:
\begin{itemize}[leftmargin=*,noitemsep]
\item Response format compliance (e.g.\ markdown vs.\ tabular output
  required by downstream systems).
\item Domain-specific factual guardrails (e.g.\ regulatory compliance
  in financial services).
\item Persona-specific tone and reading level assessment for different
  user segments.
\item Client-specific business rules and KPI definitions that do not
  appear in any public benchmark and would otherwise require manual
  review.
\end{itemize}

\subsection{Parallel Execution at Scale}

Test cases are evaluated in parallel using Python's
\texttt{ThreadPoolExecutor}. Per-row exceptions are caught and recorded
without interrupting other workers.

\subsection{Input Schema and Longitudinal Tracking}

Each evaluation run is identified by a stable \texttt{EVAL\_RUN\_ID}
and metadata (\texttt{AGENT\_NAME}, \texttt{BATCH\_ID},
\texttt{EVAL\_RUN\_DATE}), supporting longitudinal queries such as:
``How did WHERE-clause recall on hard queries change between agent v2.1
and v2.2 across the November test batch?''

\pendingedit{Note that the current implementation compares pre-captured
result snapshots; it cannot detect cases where a generated query would
have failed on the full production database due to data volume or schema
changes that occurred after snapshot creation. Capturing live execution
results directly from the production environment is a planned extension.}

\subsection{Multimodal RAG Evaluation (Work in Progress)}

\pendingedit{Integration of multimodal retrieval-augmented generation
(RAG) evaluation is currently in development. This will extend
\frameworkname{}'s response faithfulness metric to assess: (1) whether
chart and visualization outputs generated by the agent accurately reflect
the underlying data; (2) whether image-grounded claims in mixed
text-image responses are verifiable from retrieved documents. The
evaluation protocol will follow the decompose-and-verify structure of
\S\ref{sec:agentic}, adapted for multimodal evidence.}

\section{Discussion}
\label{sec:discussion}

\subsection{Connecting Design Choices to Prior Work}

Each major design choice in \frameworkname{} traces to a limitation in
the prior literature.

The choice to use an LLM for SQL component extraction is a direct
response to the evaluation gap created by the success of LLM-based SQL
generation: as \citet{gao2023} and \citet{pourreza2023dinsql} showed,
LLM-generated SQL is structurally richer than hand-authored SQL, and
this richness breaks the deterministic parsers that Spider-style
evaluation was built on \citep{yu2018spider}. Spider 2.0
\citep{lei2024spider2} recognizes the same problem and moves to workflow
evaluation. We complement this by maintaining fine-grained metric
granularity at the component level.

The hybrid execution accuracy (Hybrid-EX) design is a response to the
column-aliasing problem that neither Spider \citep{yu2018spider} nor
BIRD \citep{li2024bird} addressed, because their gold queries were
written to produce identically-named columns. Test suite accuracy
\citep{zhong2020semantic} addresses semantic equivalence through database
perturbation; \frameworkname{}'s approach addresses it through structural
metadata inference. The two approaches are complementary. Our empirical
validation demonstrates that the resulting metric achieves Substantial
human agreement ($\kappa = 0.717$), whereas the closest competing framework
(BIRD M3) achieves only $\kappa = 0.395$ (Fair).

The batched agentic evaluation design is motivated by production cost
constraints absent from academic evaluation contexts. A framework costing
\$5 per test-case evaluation run will not be used continuously;
continuous use transforms evaluation from a one-time quality gate into a
feedback loop for iterative improvement.

\subsection{Positioning Relative to Competing Frameworks}
\label{sec:positioning}

\frameworkname{} occupies a specific position in the evaluation landscape,
and we describe it here without overstating the strength of the evidence.

\paragraph{One framework among several that address unified evaluation.}
BIRD-Interact \citep{birdinteract2025} and Spider 2.0 \citep{lei2024spider2}
address overlapping territory. \frameworkname{}'s differentiator is
production deployment calibrated against human expert labels, within a
governed enterprise data environment, with a focus on the practical
failure modes that matter to enterprise practitioners. Both of those
frameworks represent strong prior work and \frameworkname{} is best
understood as complementary to them rather than as a replacement.
In concrete terms, BIRD-Interact does not address the column-aliasing
and numeric-tolerance failure modes that are routine in enterprise
LLM-generated SQL (\S\ref{sec:hybrid_ex}); it does not provide
human-calibrated execution accuracy validated against expert labels on
production data (\S\ref{sec:validation}); and it does not operate
within a governed enterprise data environment where raw results remain
under client data residency controls (\S\ref{sec:deployment}).
\frameworkname{}'s Excess Tool Usage metric (\S\ref{sec:agentic}) and
custom judge prototyping interface (\S\ref{sec:config}) also have no
direct counterpart in BIRD-Interact. These differences reflect the gap
between a research benchmark and a production evaluation pipeline.

\paragraph{An academic-benchmark reference point: FLEX.}
FLEX's published $\kappa = 0.87$ \citep{kim2025flex} was achieved on
Spider and BIRD, benchmarks in which gold queries produce
identically-named columns and numeric tolerances are not a factor; these
are precisely the failure modes that Hybrid-EX was designed to address.
The informative comparison between the two approaches would require
evaluation on enterprise data where column aliasing is prevalent; as
discussed in \S\ref{sec:related}, this test cannot currently be
conducted without non-trivial adaptation of the FLEX codebase. The two
metrics are therefore best understood as addressing different segments
of the evaluation landscape: FLEX optimizes for LLM emulation of expert
judgment on clean academic schemas, while Hybrid-EX optimizes for
calibrated, reproducible scoring on the structurally noisy SQL that
enterprise LLMs produce.

\paragraph{An evaluation approach grounded in human calibration on
enterprise data.}
The $\kappa = 0.717$ result and the pairwise baseline comparisons in
\S\ref{sec:validation} provide evidence that \frameworkname{}'s Hybrid-EX
metric reflects human expert judgment more reliably than the six
compared frameworks on the 150-query corpus used in this study.
These results are specific to the financial services and retail verticals
evaluated; broader validation across additional enterprise domains is
ongoing and necessary before generalized claims can be made. The
confidence intervals are wide (CI width 0.18--0.29) at the current
sample size, and future work will expand the annotated dataset.

\paragraph{A calibrated balance between leniency and strictness that
reflects enterprise data delivery goals.}
Enterprise AI users care fundamentally about whether the \emph{data
delivered is correct}, not whether the SQL that produced it matches a
reference query structurally. Column-name variations are routine in
LLM-generated SQL: an agent that returns \texttt{revenue} when the gold
query uses \texttt{total\_revenue} is not wrong, and an evaluation metric
that fails it is misaligned with how a data practitioner would assess the
output. As shown in \S\ref{sec:validation}, most evaluated frameworks
exhibit systematic biases in this direction: Defog M1, IBM EvalAssist M2,
BIRD M3, and RAGAS M4 have false negative rates of 42.4--45.6\%
(over-strict), penalizing queries that a domain expert would consider
correct. Dr.\ Spider M5 goes in the opposite direction, with specificity
of only 22.4\% (over-lenient). Spider2 M6 shows the inverse failure:
100\% specificity with 26.1\% sensitivity, meaning it almost never passes a
query. Hybrid-EX achieves sensitivity 81.5\% and specificity 93.1\%
(sens--spec gap 11.6pp), the most calibrated profile observed. In
practice, over-strict evaluation erodes confidence in evaluation
pipelines and may suppress deployment of systems that are correct, while
over-lenient evaluation masks real failures; the balanced operating point
of Hybrid-EX is better suited to longitudinal quality monitoring.

\paragraph{A metric with zero scoring-stage variance by design.}
Unlike stochastic LLM-based judges, Hybrid-EX's scoring stage is
deterministic by construction: given the same Stage~1 alignment
metadata, it produces identical scores for identical inputs. For
longitudinal regression testing in production environments where the
question is ``did this agent version improve over last month?'' the
scoring-stage determinism plus Stage~1 caching delivers full
reproducibility; LLM-EX cannot offer this guarantee at any stage.

\paragraph{Rapid prototyping of custom, production-grade judges and
metrics.}
A further differentiator is the ability to develop and ship
production-grade judges and metrics that are tailored to a specific
client, in addition to the standard suite reported in this paper.
Existing benchmarks fix their metric set at publication; once a
practitioner encounters a client-specific failure mode that the
benchmark does not measure, the only options are manual review or
out-of-band scripts that drift from the rest of the evaluation. In
\frameworkname{}, custom judges and metrics are first-class
citizens of the same pipeline: they are prototyped against the existing
golden test set, validated against human labels where those exist,
and surfaced in the same longitudinal report as the standard metrics.
This shortens the path from ``we noticed a recurring issue specific to
this client'' to ``we have a reproducible, governed metric for it,''
and it lets the standard suite act as a stable backbone while
client-specific evaluation grows on top of it.

\subsection{Future Directions}
\label{sec:future}

Natural next steps include: expanding the annotated dataset to narrow
the $\kappa$ confidence intervals (currently 0.18--0.29 width at $n =
150$); extending the agentic suite to evaluate chart and visualization
outputs (currently in development, \S\ref{sec:deployment}); integration
with online A/B testing infrastructure for real-time quality monitoring;
broader enterprise domain validation across manufacturing, logistics, and
life sciences; and a composite enterprise AI quality score that weights
metrics by business impact for a given deployment context. The current
tool-order metric assumes a unique correct ordering; a DAG-based gold
specification representing valid orderings as a partial order is a
natural extension that would support more complex multi-step agent
workflows.

\section{Acknowledgments}
\label{sec:acknowledgments}
The authors thank \textbf{Corey Koberg} for his thoughtful feedback on
earlier drafts of this work, and \textbf{Balamurugan Murugesan} for
his contributions during the development of LLM judges \frameworkname{}. We also
thank the annotation team whose careful labeling made the
human-calibration study possible, and our enterprise clients across
financial services and retail whose deployments motivated the design
choices described here. \frameworkname{} is developed within
Merkle Analytics.
\section{Conclusion}
\label{sec:conclusion}

We have presented \frameworkname{}, \orgname{}'s unified evaluation framework
for enterprise AI systems that combine text-to-SQL generation with
agentic tool orchestration. \frameworkname{} makes three primary
contributions. LLM-assisted SQL component extraction extends Spider-style
evaluation to the SQL that LLMs actually produce in production.
Hybrid execution accuracy (Hybrid-EX), our primary contribution, bridges
the gap between LLM-based and execution-based evaluation by using an LLM
to resolve structural ambiguities before applying fully deterministic
cell-level scoring. Validated against 150 human-annotated industry
queries, it achieves Cohen's $\kappa = 0.717$ [95\% CI: 0.600--0.822]
(Substantial agreement per Landis \& Koch, 1977), a balanced accuracy of
87.3\%, and a sensitivity--specificity gap of 11.6pp, significantly
outperforming all six evaluated baselines ($\Delta\kappa$ range:
0.322--0.502, all $p \leq 0.001$). The enterprise agentic evaluation
suite integrates established metrics from RAGAS, G-Eval, and prior agent
benchmark work into a unified production evaluation pipeline; the Excess
Tool Usage metric is its sole novel element, and no claim is made
regarding the individual agentic metrics beyond their utility in an
integrated production harness.

\frameworkname{} is delivered as an internal evaluation tool that
runs entirely within the client's governed data environment, enabling
\orgname{} clients to run evaluations without their data leaving that
environment. The same tool supports configurable LLM judge backends
and the rapid prototyping of production-grade, client-specific judges
and metrics alongside the standard suite. We release the full
methodology here with the aim of raising evaluation standards across the
field.

\frameworkname{}'s design reflects specific choices about what matters in
enterprise deployments, and we summarize its positioning here for readers
approaching the paper from its conclusion. Among the evaluated frameworks,
it is distinguished by three characteristics. First, it is calibrated
against human expert labels on real enterprise data, with the validation
results specific to the financial services and retail corpora evaluated;
broader validation is ongoing. Second, its execution accuracy metric is
designed to handle the column-name variations, numeric tolerances, and
ordering ambiguities that are routine in LLM-generated SQL. These are failure
modes that existing benchmarks did not address because their gold queries
were written to produce identically-named columns. The empirical
consequence is that competing frameworks reject 42--46\% of queries that
human experts consider correct; Hybrid-EX reduces this false negative rate
substantially while maintaining high specificity (93.1\%). Third, its
scoring stage is deterministic by construction, producing identical
scores for identical inputs once Stage~1 alignment metadata is fixed
(or cached), a property that is essential for longitudinal quality
monitoring in production environments. This is the kind of continuous feedback
loop through which enterprise AI systems actually improve. Taken
together, these characteristics make \frameworkname{} suited to serve
as a continuous evaluation backbone rather than a one-time quality gate.

\bibliographystyle{plainnat}
\bibliography{references}

\appendix

\section{Prompt Design Details}

\subsection{SQL Component Extraction Prompt}

The system prompt establishes the model as an expert SQL parser
implementing Spider evaluation methodology. The user prompt defines all
seven component types with examples, emphasizing lowercase normalization,
schema-prefix stripping, CTE transparency (extract from the final SELECT,
not the CTE body), and aggregation function enumeration. Temperature is
set to $T = 0.0$.

\subsection{SQL LLM Judge Prompt}

The system prompt establishes the evaluation goal as semantic
correctness, not syntactic similarity. The user prompt provides the user
question, agent interpretation, gold SQL, and generated SQL, then
specifies the 0--10 scoring rubric and the explicit list of patterns that
are acceptable (CTEs, positional GROUP BY, double-underscore aliases,
schema qualification, LEFT vs.\ INNER JOIN when results are identical,
NULL-handling helpers). Temperature is set to $T = 0.1$.

\subsection{Output Comparison Prompt}

The system prompt establishes the model as a data comparison analyst.
The user prompt provides both queries and result tables (up to 100 rows
in CSV format) and requests ten structured output fields: matching
category, judge score, column mapping, trivial columns, sort keys,
common/unique columns, data type annotations, and deterministic
evaluation hints. Temperature is set to $T = 0.1$.

\subsection{Combined Agentic Evaluation Prompt}

The system prompt establishes simultaneous evaluation across faithfulness,
G-Eval, and intent resolution. A structured JSON output schema with
strict property typing is enforced. Temperature is set to $T = 0.1$.

\end{document}